# Suboptimality Bounds for Stochastic Shortest Path Problems

Eric A. Hansen
Dept. of Computer Science and Engineering
Mississippi State University
Mississippi State, MS 39762
hansen@cse.msstate.edu

## Abstract

We consider how to use the Bellman residual of the dynamic programming operator to compute suboptimality bounds for solutions to stochastic shortest path problems. Such bounds have been previously established only in the special case that "all policies are proper," in which case the dynamic programming operator is known to be a contraction, and have been shown to be easily computable only in the more limited special case of discounting. Under the condition that transition costs are positive, we show that suboptimality bounds can be easily computed even when not all policies are proper. In the general case when there are no restrictions on transition costs, the analysis is more complex. But we present preliminary results that show such bounds are possible.

## 1 Introduction

A stochastic shortest path problem is a Markov decision process (MDP) where the objective is to find a minimum-cost policy that reaches a goal or terminal state with probability 1 [4, 2]. It is an elegant model for many problems of planning under uncertainty, especially for goal-oriented decision-theoretic planning problems where policy execution terminates once a goal condition is achieved. Standard solution methods rely on dynamic programming or linear programming. The model is also used in the development and analysis of reinforcement learning and heuristic search algorithms for MDPs [15, 1, 6]. There are extensions of the stochastic shortest path problem for planning under partial observability [11], multi-agent planning [12, 8], and risk-sensitive planning [10].

For the stochastic shortest path problem, the expected total cost of policy execution is bounded, without discounting. Thus it is an important alternative to the discounted infinite-horizon MDP as a model for decision-theoretic planning.

Although use of a discount factor sometimes has an economic justification, discounting is not well-motivated for many AI planning problems and has potential drawbacks. It can skew the relative values of policies and change the optimal policy. Discounting can also make it impossible to guarantee that dynamic programming finds a policy that reaches a goal state with probability 1. With discounting, a policy that cycles forever without reaching the goal state still has finite total cost, which could be less than the cost of the best policy that is guaranteed to reach the goal state.

Despite potential drawbacks, the discounted infinite-horizon model is widely used. One reason for its appeal is that discounting is a simple way to ensure that algorithms for solving infinite-horizon MDPs have the desired convergence properties, since the dynamic programming operator is a contraction operator in this case, with the contraction rate equal to the discount factor. Although the convergence of value iteration and policy iteration for stochastic shortest path problems is well-established, it is not based on a contraction property (except in the special case that all policies are proper). As a result, there is no guarantee that the convergence rate is geometric and suboptimality bounds are generally not available for solutions found by dynamic programming, in contrast to the discounted case.

In this paper, we show that even though the dynamic programming operator for stochastic shortest path problems is not a contraction operator in general, it behaves like a contraction operator if the dynamic programming algorithm is started with the value function of a *proper policy*, which is a policy that achieves the goal state with probability one. Under the additional condition that transition costs are positive, we show how to use the Bellman residual of the dynamic programming operator to compute suboptimality bounds. In the general case where there are no restrictions on action costs, the analysis is more complex. But we establish some preliminary results that support a general approach to computing bounds. Our results apply to both completely observable and partially observable stochastic shortest path problems.

## 2 Background

We begin with a review of the stochastic shortest path problem as formulated by Bertsekas and Tsitsiklis [3, 4, 5] and extended to the partially observable case by Patek [11]. We also review previous work on computing suboptimality bounds for solutions found by dynamic programming.

### 2.1 Stochastic shortest path problem

Like any discrete-time Markov decision process (MDP), a stochastic shortest path problem includes a set of states, $S$, and a set of control actions, $U$, which we assume are both finite; a set of transition probabilities, where $p_{ij}(u)$ denotes the probability that the system moves to state $j \in S$ after action $u \in U$ is taken in state $i \in S$; and a set of real-valued costs, where $g(i, u, j)$ denotes the cost incurred when action $u$ taken in state $i$ results in a transition to state $j$. The expected cost of taking action $u$ in state $i$ is denoted $g(i, u) = \sum_{j \in S} p_{ij}(u) g(i, u, j)$.

In addition, a stochastic shortest path problem is characterized by a set of assumptions from which it derives its special properties. The first is the following.

**Assumption 1.** *The state set includes a special* target *or* terminal state, $t \in S$, *which is zero-cost and absorbing, which means that* $p_{tt}(u) = 1$ *and* $g(t, u, t) = 0, \forall u \in U$.

The objective is to find a *stationary policy*, $\mu : S \to U$, that reaches the terminal state while minimizing the total expected cost. We are especially interested in stationary policies that reach the terminal state with probability 1 from any initial state, called *proper policies*.

**Definition 1** (Bertsekas and Tsitsiklis). *A stationary policy $\mu$ is said to be* proper *if there exists a finite positive integer $m$ such that there is a positive probability of reaching the terminal state after at most $m$ stages when following this policy, regardless of the initial state, that is,*

$$\rho_\mu = \min_{i \in S} P(x_m = t | x_0 = i, \mu) > 0, \quad (1)$$

*where $x_k$ denotes the state of the process at stage $k$. A stationary policy that is not proper is said to be* improper.

Following a policy that is proper according to this definition, a process reaches the terminal state with probability 1, regardless of the initial state [2].

The *value (or cost-to-go) function* $J_\mu$ of a stationary policy $\mu$ gives the expected total cost of following the policy starting from an initial state $i$, defined as

$$J_\mu(i) = \lim_{N \to \infty} E_\mu \left[ \sum_{k=0}^{N-1} g(x_k, \mu(x_k)) | x_0 = i \right], \forall i \in S. \quad (2)$$

For a proper policy $\mu$, the expected cost is bounded above for each state, that is, $J_\mu(i) < \infty, \forall i \in S$.

A stochastic shortest path problem is solved by finding an optimal policy $\mu^*$ satisfying

$$J^*(i) = J_{\mu^*}(i) \leq J_\mu(i), \forall i \in S, \forall \mu \in M, \quad (3)$$

where $J^*$ denotes the optimal value function. Two additional assumptions of the stochastic shortest path problem ensure that an optimal policy exists and that it is proper.

**Assumption 2.** *There exists at least one proper policy.*

**Assumption 3.** *For every improper policy $\mu$, the corresponding cost $J_\mu(i)$ is infinite for at least one state $i$.*

Assumption 3 is equivalent to the assumption that the expected cost $J_\mu(i)$ of state $i$ under policy $\mu$ is infinite if a process started in state $i$ and following policy $\mu$ does not reach the terminal state with probability 1.

Two special cases of the stochastic shortest path problem play an important role in the analysis of Bertsekas and Tsitsiklis [3, 4, 5]. The first is the stochastic shortest path problem when *all* policies are proper. Analysis of this special case is easier, although the assumption that all policies are proper is often unrealistic. Another special case is the discounted infinite-horizon MDP. By a well-known reduction, any discounted infinite-horizon MDP can be reduced to an equivalent stochastic shortest path problem in which, for every state and action pair, there is a probability $(1 - \beta)$ of making a transition to the terminal state, where $\beta < 1$ is the discount factor, with the other transition probabilities normalized. For the stochastic shortest path problem created by this reduction, all policies are proper. As we will see, the well-known convergence properties and suboptimality bounds for the discounted infinite-horizon MDP are a special case of those for the stochastic shortest path problem.

### 2.2 Dynamic programming

The stochastic shortest path problem can be solved using dynamic programming, where the dynamic programming operator is defined as follows,

$$TJ(i) = \min_{u \in U} \sum_{j \in S} p_{ij}(u) \left( g(i, u, j) + J(j) \right), \forall i \in S, \quad (4)$$

and the related policy evaluation operator $T_\mu$ is defined as,

$$T_\mu J(i) = \sum_{j \in S} p_{ij}(\mu(i)) \left( g(i, \mu(i), j) + J(j) \right), \forall i \in S. \quad (5)$$

These operators are the key steps in two dynamic programming algorithms: value iteration and policy iteration.

Analysis of the convergence of value and policy iteration turns on two properties of the dynamic programming and policy evaluation operators: the monotonicity property and the contraction property. The *monotonicity property* is defined as follows: if $J \leq J'$, then $TJ \leq TJ'$ and $T_\mu J \leq T_\mu J'$.[1] Both operators satisfy the monotonicity property.

---
[1] As shorthand, we let $J \leq J'$ denote $J(i) \leq J'(i), \forall i \in S$.

An operator $T$, such as the dynamic programming operator, is said to be a *contraction operator* if, for all bounded value functions $J$ and $J'$,

$$||TJ - TJ'|| \leq \beta||J - J'||, \qquad (6)$$

where $\beta$, with $0 \leq \beta < 1$, is the contraction rate and $||.||$ is some norm. For discounted infinite-horizon MDPs, it is well-known that both the dynamic programming operator $T$ and the policy evaluation operator $T_\mu$ are contraction operators, with the contraction rate $\beta$ equal to the discount factor. If the state space is finite, they are contraction operators in the maximum norm, defined as $||x|| = \max_{i \in S} x(i)$. If the state space is continuous, they are contraction operators in the supremum norm, defined as $||x|| = \sup_{i \in S} x(i)$. For MDPs, the maximum or supremum norm, $||TJ - J||$, is called the *Bellman residual*.

For the stochastic shortest path problem, there is no discount factor (or equivalently, the discount factor is 1). In this case, a related concept of contraction operator is useful. An operator $T$ is said to be an *m-stage contraction operator* if $T^m$, which is the composition of $T$ with itself $m$ times, is a contraction operator, that is, if there is some $\beta$, with $0 \leq \beta < 1$, and some norm $||.||$, such that

$$||T^m J - T^m J'|| < \beta ||J - J'||, \qquad (7)$$

for all bounded value functions $J$ and $J'$. For the stochastic shortest path problem, the dynamic programming operator is an $m$-stage contraction operator in the maximum (or supremum) norm if there is some minimum positive probability $\rho_m = (1 - \beta) > 0$ that the process reaches the terminal state within $m$ steps, beginning from any state. Bertsekas and Tsitsiklis [3, 4, 5] show that the dynamic programming operator for stochastic shortest path problems is an $m$-stage contraction operator in the special case that all policies are proper.[2] As for the policy evaluation operator, it is an $m$-stage contraction if it evaluates a proper policy.

The significance of showing that the dynamic programming and policy evaluation operators satisfy the contraction property, under the condition that all policies are proper, is that the convergence of both value iteration and policy iteration follows from Banach's Fixed-Point Theorem; it also follows that the convergence rate is geometric. Bertsekas and Tsitsiklis consider this special case first because the dynamic programming operator is not a contraction operator in the general case. In the words of Bertsekas and Tsitsiklis [4], "our assumptions do not imply that the corresponding dynamic programming mapping is a contraction (unlike the situation in discounted problems), unless all policies are proper." They give an example to show that the contraction property does not hold in general. It is a stochastic shortest path problem with two states: state 1 is terminal and state 2 is nonterminal. Two actions are possible in the nonterminal state. One causes a deterministic transition to the terminal state with a cost of 2; the other causes a deterministic self-transition with a cost of 1. Thus there are two stationary policies, one proper and the other improper. Consider two value functions $J$ and $J'$ for which $J(1) = J'(1) = 0$, $J(2) < 1$, and $J'(2) < 1$. Under these conditions, $|TJ(2) - TJ'(2)| = |(J(2) + 1) - (J'(2) + 1)| = |J(2) - J'(2)|$, which shows that, in general, $T$ is not a contraction with respect to any norm.

The principal contribution of Bertsekas and Tsitsiklis to the theory of stochastic shortest path problems is their proofs of the convergence of value iteration and policy iteration in the general case, where the dynamic programming operator is not a contraction. Given only the assumptions of the stochastic shortest path problem, they prove the following.

- The optimal value function $J^*$ is the unique solution of the Bellman equation $J^* = TJ^*$, and *value iteration* converges to it, which means that for any initial value function $J$, $\lim_{t \to \infty} T^t J = J^*$.

- The value function $J_\mu$ of a proper policy $\mu$ is the unique solution of the linear system of equations, $J_\mu = T_\mu J_\mu$, and *policy evaluation* converges to it, which means $\lim_{t \to \infty} T_\mu^t J = J_\mu$, for any initial $J$.

- Starting with a proper policy $\mu^0$, *policy iteration* generates a sequence of policies $\mu^1, \mu^2, \ldots$, for which $J_{\mu^{k+1}} \leq J_{\mu^k}, \forall k$, by alternating a policy evaluation step that computes $J_{\mu^k}$, with a *policy improvement* step that computes an improved policy $\mu^{k+1}$ using the equation $T_{\mu^{k+1}} J_{\mu^k} = T J_{\mu^k}$, or equivalently, $\forall i \in S$,

$$\mu^{k+1}(i) = \arg\min_{u \in U} \sum_{j \in S} p_{ij}(u)\big(g(i,u,j) + J_{\mu^k}(j)\big), \qquad (8)$$

and the sequence converges to an optimal policy.

However, their convergence results are weaker than those based on the contraction property and Banach's Fixed-Point Theorem because they establish pointwise convergence, not uniform convergence. As a result, they provide no guarantee that the rate of convergence is geometric and no way to bound the number of iterations until convergence to an $\epsilon$-optimal policy. They also do not allow easy use of the Bellman residual to compute suboptimality bounds.

---

[2] The minimum probability $\rho_m$ that the terminal state will be reached in $m$ stages can be computed by solving a finite-horizon MDP with the same transition probabilities as the stochastic shortest path problem but with a cost of 1 for any transition to the terminal state, and a cost of 0 for all other transitions. Finding a policy that minimizes the $m$-horizon cost corresponds to finding a policy that minimizes the probability of reaching the terminal state within $m$ stages. By Definition 1, the minimum probability of reaching the terminal state within $m$ stages is greater than zero, for some positive integer $m$, if all policies are proper.

## 2.3 Suboptimality bounds

We next review how to use the Bellman residual of the dynamic programming operator to bound the suboptimality of solutions found by dynamic programming. For stochastic shortest path problems, Bertsekas [2, pp. 413–414] gives the following result.

**Theorem 1** (Bertsekas). *For any stochastic shortest path problem, any value function $J$, a greedy policy $\mu$ with respect to $J$, and for all $i \in S$, the following bounds hold,*

$$J(i) + \underline{c}N^*(i) \leq J^*(i) \leq J_\mu(i) \leq J(i) + \overline{c}N_\mu(i), \quad (9)$$

*where $\underline{c} = \min_{i \in S}[TJ(i) - J(i)]$, $N^*(i)$ is an upper bound on the expected number of steps needed to reach the terminal state $t$ beginning from state $i$ and following an optimal policy, $\overline{c} = \max_{i \in S}[TJ(i) - J(i)]$, and $N_\mu(i)$ is an upper bound on the expected number of steps needed to reach $t$ beginning from state $i$ and following policy $\mu$.*

Some simplifications can help bring this result into focus. Leaving out the inequalities involving $J_\mu$ and subtracting $J(i)$, we have: $\underline{c}N^*(i) \leq J^*(i) - J(i) \leq \overline{c}N_\mu(i)$. Note that $\underline{c} \leq 0$ and $\overline{c} \geq 0$. Note also that the Bellman residual is: $||TJ - J|| = \max\{-\underline{c}, \overline{c}\}$. Thus we have:

$$|J^*(i) - J(i)| \leq ||TJ - J|| \cdot \max\{N^*(i), N_\mu(i)\}. \quad (10)$$

Since by definition, $||J^* - J|| = \max_{i \in S}|J^*(i) - J(i)|$, we finally have:

$$||J^* - J|| \leq ||TJ - J|| \cdot \max_{i \in S}\max\{N^*(i), N_\mu(i)\}. \quad (11)$$

In addition to this bound on the suboptimality of a value function $J$, the inequalities $J_\mu(i) \leq J(i) + \overline{c}N_\mu(i)$ in Equation 9 let us bound the suboptimality of a greedy policy $\mu$ with respect to $J$. It follows that we can compute suboptimality bounds if we can compute the bounds $N^*(i)$ and $N_\mu(i)$ on the expected number of stages until termination. "Unfortunately," writes Bertsekas [2, pp. 413–414], these bounds "are easily computed or approximated only in the presence of special problem structure."

Bertsekas mentions just one special case in which these bounds can be easily computed: the discounted infinite-horizon case. As already pointed out, any discounted infinite-horizon MDP (with discount factor $\beta$) can be reduced to an equivalent stochastic shortest path problem in which, for every state and action pair, there is a probability $(1 - \beta)$ of making a transition to a terminal state, with the other transition probabilities normalized accordingly. It follows that the expected number of steps until termination, from any starting state, is $\sum_{t=0}^{\infty} \beta^t = 1/(1 - \beta)$. Letting $N^*(i) = N_\mu(i) = 1/(1 - \beta), \forall i \in S$, the well-known bound on the suboptimality of a value function $J$, which is $||J^* - J|| \leq ||TJ - J||/(1 - \beta)$, is seen to be a special case of the suboptimality bound given by Equation 11.

Bertsekas and Tsitsiklis [5, pp. 23–24] describe how to bound the expected number of stages until termination for any stochastic shortest path problem for which all policies are proper, although it requires solving an MDP of the same size as the original stochastic shortest path problem. Given a stochastic shortest-path problem, consider a related infinite-horizon MDP where the transition probabilities are the same but there is a cost of $0$ for any transition to the terminal state and all other transitions incur a cost of $-1$. For this MDP, finding a policy that minimizes the expected infinite-horizon cost corresponds to finding a policy that maximizes the *expected* number of stages it takes to reach the terminal state. The values computed by solving this MDP bound the number of stages until termination for *any* policy. Obviously, these bounds are finite (and the MDP is well-defined) if and only if all policies are proper.

If a greedy policy $\mu$ with respect to value function $J$ is not proper, then $N_\mu(i)$ is not finite for at least one state $i$ and the bounds of Equations 10 and 11 are not finite. We address the challenge of how to compute bounds when not all policies are proper beginning in Section 3.

## 2.4 Partial observability

Patek [11] extends the framework of stochastic shortest path problems to the partially observable case. A partially observable MDP (POMDP) includes the same states, actions, transition probabilities and costs defined earlier, plus a finite set of observation symbols, $Z$, and a set of observation probabilities, where $p_z(j, u)$ denotes the probability that symbol $z \in Z$ is observed after action $u \in U$ results in a transition to state $j \in S$. In addition to the three assumptions of a stochastic shortest path problem given in Section 2.1, a partially observable stochastic shortest path problem includes an assumption that ensures that termination of the process is perfectly recognized.

**Assumption 4.** *The set of observation symbols includes a special symbol, $z_t \in Z$, which is unique to transitions to the terminal state $t$. That is, $p_{z_t}(t, u) = 1$ and $p_{z_t}(j, u) = 0, \forall u \in U, j \in S$.*

As is well-known, a POMDP can be solved by dynamic programming if it is first transformed into an equivalent completely observable MDP over belief states, where a belief state is an $|\mathcal{S}|$-dimensional vector of probabilities maintained by Bayesian conditioning. Given the assumptions of the partially observable stochastic shortest path problem, Patek [11] shows that value iteration and policy iteration have the same convergence properties established by Bertsekas and Tsitsiklis [3, 4] in the completely observable case. His analysis follows the same outline. In the special case that all policies are proper, he shows that the dynamic programming operator is an $m$-stage contraction operator. In the general case when not all policies are proper, he proves convergence without using a contraction property.

## 3 Uniform improvability and proper policies

The convergence proofs for value iteration and policy iteration given by Bertsekas and Tsitsiklis [3, 4] and Patek [11] for the general case when not all policies are proper are significant because they do not depend on the contraction property. But without a contraction property, they do not ensure a geometric rate of convergence or provide a way to use the Bellman residual to compute suboptimality bounds. In the rest of the paper, we show a way around this.

First, we establish a condition under which a policy found by dynamic programming is guaranteed to be proper.

**Theorem 2.** *For any stochastic shortest path problem and any value function $J$ for which $TJ \leq J$, a greedy policy $\mu$ with respect to $J$, defined as*

$$\mu(i) = \arg\min_{u \in U} \sum_{j \in S} p_{ij}(u)(g(i,u,j) + J(j)), \forall i \in S, \quad (12)$$

*is a proper policy and $J_\mu \leq J$.*

*Proof.* The key observation is that $TJ = T_\mu J$, which means that application of the dynamic programming operator $T$ can be viewed as the first application of the policy evaluation operator $T_\mu$ in evaluating a greedy policy $\mu$ with respect to $J$. When $TJ \leq J$, we have $T_\mu J \leq J$. By the monotonicity property, it follows that every successive iteration of $T_\mu$ monotonically improves the value function, and thus $J_\mu = \lim_{n \to \infty} T_\mu^n J \leq T_\mu J = TJ \leq J$. Since $J_\mu$ is bounded above for every state, $\mu$ must be proper. $\square$

We call a value function $J$ for which $TJ \leq J$ a *uniformly improvable* value function, a term used by others [16, 14]. Consider the subspace of uniformly improvable value functions: $\mathcal{J} = \{J | TJ \leq J\}$. By the monotonicity property, this subspace is closed under the dynamic programming operator. It follows that if value iteration is started with a value function $J \in \mathcal{J}$, a greedy policy with respect to this value function, and a greedy policy with respect to *any* subsequent value function found after any number of iterations of value iteration, must be proper. We already have the same guarantee for policy iteration. The convergence proof for policy iteration given by Bertsekas and Tsitsiklis (and by Patek in the partially observable case) requires the initial policy to be proper; otherwise, the value function computed by policy evaluation is not bounded. Given an initial proper policy, it follows from the policy improvement theorem that any policy found after any number of iterations of policy iteration must be proper, since the cost of an improved policy cannot increase for any state. Indeed, this guarantee holds precisely because the value function of an initial proper policy $\mu$ is uniformly improvable; note that $TJ_\mu = J_\mu$ implies $TJ_\mu \leq J_\mu$. It follows that whether we use policy iteration or value iteration, we can guarantee uniform improvability by finding an initial proper policy.

Different algorithms can be used to find an initial proper policy. Since the better the initial policy and value function, the sooner policy iteration or value iteration converges, extra computational effort spent trying to find an initial proper policy that is of high quality could be well-spent. However, it is not difficult to find *some* initial proper policy.

**Theorem 3.** *For any stochastic shortest path problem, the uniform random policy is proper.*[3]

*Proof.* By Assumption 2, there exists some proper policy $\mu$. By Definition 1, there is a positive integer $m$ such that after $m$ stages, there is a probability $\rho_\mu > 0$ that following policy $\mu$ leads to the terminal state. Since a uniform random policy selects an action at random based on a uniform probability distribution, it executes the same action as policy $\mu$ for $m$ consecutive stages with probability $(1/|U|)^m > 0$. It follows that the probability of reaching the terminal state within $m$ stages by following the uniform random policy is greater than or equal to $(1/|U|)^m \cdot \rho_\mu$, which is positive, and thus the uniform random policy is proper. $\square$

When the value function is uniformly improvable, we can simplify the bounds given in Theorem 1 and Equations 10 and 11. Note that when $TJ \leq J$, we have $\bar{c} = 0$. From Equation 9, it follows that $J^*(i) \leq J_\mu(i) \leq J(i)$, where $\mu$ is a greedy policy with respect to $J$. Thus $\mu$ must be proper. (This is the same result proved in Theorem 2.) Setting aside the bounds involving $J_\mu$, consider the remaining bounds: $J(i) + \underline{c}N^*(i) \leq J^*(i) \leq J(i)$. Subtracting $J(i)$, we have: $\underline{c}N^*(i) \leq J^*(i) - J(i) \leq 0$. Thus $|J^*(i) - J(i)| \leq -\underline{c}N^*(i)$. By assumption, $TJ \leq J$, and so the Bellman residual is: $||TJ - J|| = -\underline{c}$. Thus we have,

$$|J^*(i) - J(i)| \leq ||TJ - J|| \cdot N^*(i), \quad (13)$$

where $N^*(i)$, as defined in Theorem 1, is an upper bound on the expected number of steps needed to reach the terminal state beginning from state $i$ and following an optimal policy. By definition, $||J^* - J|| = \max_{i \in S} |J^*(i) - J(i)|$, and so,

$$||J^* - J|| \leq ||TJ - J|| \cdot N^*, \quad (14)$$

where $N^* = \max_{i \in S} N^*(i)$ is an upper bound on the expected number of steps needed to reach the terminal state from *any* other state by following an optimal policy. Note that Equation 13 is a simplification of Equation 10 and Equation 14 is a simplification of Equation 11, where both simplifications are possible because the value function $J$ is uniformly improvable. The condition that $J$ is uniformly improvable also allows the following simplification.

**Theorem 4.** *If value function $J$ is uniformly improvable, then any bound on its suboptimality is also a bound on the suboptimality of a greedy policy $\mu$ with respect to $J$.*

*Proof.* Immediate since $J_\mu(i) \leq J(i)$ if $TJ \leq J(i)$. $\square$

---
[3] I am grateful to Bruno Scherrer for the observation expressed in this theorem and the idea of the proof.

# 4 Positive transition costs

In this section, we consider a special case of the stochastic shortest path problem where all transition costs incur a positive cost, except possibly for transitions into the terminal state. By Assumption 1, a transition from the terminal state to itself has a cost of zero. Because a transition from a nonterminal state to the terminal state occurs only once, we do not need to place any restriction on its cost (except, of course, that it is bounded).

Under this condition on transition costs, we show how to compute upper bounds, denoted $N(i), \forall i \in S$, on the expected number of stages until termination of any policy $\mu$ for which $J_\mu \leq J$, where $J$ is uniformly improvable. We use the notation $N(i)$ instead of $N^*(i)$ because these bounds on the expected number of stages until termination apply to any policy $\mu$ for which $J_\mu \leq J$, not just an optimal policy. Since $J^* \leq J_\mu \leq J$, we have $N(i) \geq N^*(i), \forall i \in S$, and so we can use $N(i)$ in place of $N^*(i)$ in Equations 13 and 14 to bound the suboptimality of solutions found by dynamic programming.

The proof strategy we adopt to establish these bounds is to show that any policy $\mu$ that does not terminate within $N(i)$ stages on average, beginning from state $i$, must have an expected cost $J_\mu(i)$ greater than $J(i)$, which contradicts the assumption that $J_\mu \leq J$. The significance of this strategy is that it does not require all policies to be proper. It simply requires a uniformly improvable value function $J$.

**Theorem 5.** *For any stochastic shortest-path problem for which $g(i, u, j) > 0$ for all $i \in S, u \in U, j \in S \setminus t$, and for any policy $\mu$ with value function $J_\mu \leq J$, an upper bound on the mean number of steps until the terminal state is reached beginning from any state $i$ is*

$$N(i) = \frac{J(i) - a}{b} + 1, \quad (15)$$

*where $a = \min\{g(i, u, t) : i \in S, u \in U\}$ denotes the minimum cost of any transition into the terminal state and $b = \min\{g(i, u, j) : i \in S, u \in U, j \neq t\}$ denotes the minimum cost of any other transition.*

*Proof.* For any policy $\mu$, let $n_\mu(i)$ denote the expected number of steps until the terminal state is reached beginning from state $i$ and following policy $\mu$. Because $a$ is the minimum cost of any transition into the terminal state and $b$ is the minimum cost of any other transition, $a + b(n_\mu(i) - 1)$ is the minimum cost of any sequence of $n_\mu(i)$ transitions that ends in the terminal state; therefore, $J_\mu(i) \geq a + b(n_\mu(i) - 1)$. Now if $N(i)$ is not an upper bound on the expected number of stages until termination for some state $i$, there must be some policy $\mu$ for which both $n_\mu(i) > N(i)$ and $J_\mu(i) \leq J(i)$. But since $N(i) = (J(i) - a)/b + 1$, then $n_\mu(i) > N(i)$ implies that $J_\mu(i) > J(i)$, which contradicts the assumption that $J_\mu \leq J$. It follows that $n_\mu(i) \leq N(i), \forall i \in S$. □

Figure 1: Gridworld navigation example [13, 9].

Note that the "+1" in Equation 15 counts the transition into the terminal state. This approach to computing suboptimality bounds works best if transition costs are uniform as well as positive. If transition costs are positive but non-uniform, the bounds are still valid, but potentially looser. In this case, the bounds could be improved by considering the minimum *expected* transition cost after an action, instead of simply the minimum transition cost.

**Example** For illustration, consider the small gridworld navigation problem shown in Figure 1. Russell and Norvig [13] describe a completely observable version of this gridworld and Parr and Russell [9] describe a partially observable version. To allow reference to individual states, we number each cell of the grid from 0 to 10. Figure 1 shows all of the numbers except two; the +1 state is numbered 3 and the −1 state is numbered 6. Any action taken in either the +1 or the −1 state results in a deterministic transition to a terminal state (which is not shown) and a reward of +1 or −1 respectively. In all other states, any of the four possible navigation actions (with one corresponding to each direction of the compass) incurs a negative reward of −0.04, which is equivalent to a positive cost. (For convenience, we keep the reward-maximization framework used by Russell and Norvig [13] and Parr and Russell [9]. Note that it is easily transformed to the cost-minimization framework of a stochastic shortest-path problem.) We use the same transition and observation probabilities given by Russell and Norvig [13] and Parr and Russell [9]. For this example, it is *not* the case that all policies are proper.

One reason for adopting this simple example is that it is the same example used by Russell and Norvig [13] to illustrate how to compute suboptimality bounds for solutions found by dynamic programming for *discounted* infinite-horizon MDPs. However, this example is most naturally formalized in an *undiscounted* reward-maximization framework that is equivalent to a stochastic shortest path problem, as noted by both Russell and Norvig [13] and Parr and Russell [9]. Although this gridworld example is very simple, it helps to illustrate several aspects of our approach.

We implemented our approach to computing suboptimality bounds in exact value iteration and policy iteration algorithms for completely observable and partially observable stochastic shortest path problems. Table 1 shows the results for the first 12 iterations of the algorithms, starting from the value function of the uniform random policy. The bounds depend not only on the size of the Bellman residual, but on the upper bound $m = \max_{i \in S} N(i)$ on the expected number of stages before termination, which in turn depends on the current value function. In the first couple iterations, the error bounds are loose because the value function is still rather poor and $m$ is relatively large. As the value function improves over successive iterations, the bounds improve due to a reduction in the expected number of stages until termination, as well as a decrease in the Bellman residual.

|   | Completely observable |||||||| Partially observable ||||||||
|   | Value iteration |||| Policy iteration |||| Value iteration |||| Policy iteration ||||
|   | $\underline{J}$ | m | resid. | error | $\underline{J}$ | m | resid. | error | $\underline{J}$ | m | resid. | error | $\underline{J}$ | m | resid. | error |
|---|---|---|---|---|---|---|---|---|---|---|---|---|---|---|---|---|
| 0 | -1.603 | 66.1 | - | - | -1.603 | 66.1 | - | - | -1.603 | 66.1 | - | - | -1.603 | 66.1 | - | - |
| 1 | -1.570 | 65.3 | 0.9567 | 62.428 | -0.885 | 48.1 | 0.9567 | 46.030 | -1.603 | 66.1 | 0.9567 | 63.200 | -1.033 | 51.8 | 0.9567 | 49.570 |
| 2 | -1.430 | 61.7 | 0.8470 | 52.302 | 0.369 | 16.8 | 1.0070 | 16.880 | -1.570 | 65.3 | 0.8470 | 55.270 | 0.332 | 17.7 | 0.8828 | 18.090 |
| 3 | -1.206 | 56.1 | 0.7379 | 41.433 | 0.388 | 16.3 | 0.0186 | 0.304 | -1.430 | 61.8 | 0.7379 | 45.560 | 0.350 | 17.2 | 0.1073 | 1.851 |
| 4 | -0.876 | 47.9 | 0.6585 | 31.551 | 0.388 | 16.3 | 0.0000 | 0.000 | -1.209 | 56.2 | 0.6585 | 37.030 | 0.363 | 16.9 | 0.4136 | 0.701 |
| 5 | -0.256 | 32.4 | 0.6204 | 20.102 |   |   |   |   | -0.882 | 48.1 | 0.6165 | 29.630 | 0.374 | 16.6 | 0.0213 | 0.355 |
| 6 | 0.153 | 22.2 | 0.4094 | 9.075 |   |   |   |   | -0.266 | 32.7 | 0.4030 | 13.160 | 0.379 | 16.5 | 0.0120 | 0.204 |
| 7 | 0.263 | 19.4 | 0.2568 | 4.991 |   |   |   |   | 0.101 | 23.5 | 0.2603 | 6.110 | 0.383 | 16.4 | 0.0049 | 0.082 |
| 8 | 0.310 | 18.2 | 0.1389 | 2.534 |   |   |   |   | 0.249 | 19.8 | 0.1407 | 2.782 | 0.384 | 16.4 | 0.0011 | 0.019 |
| 9 | 0.333 | 17.7 | 0.0726 | 1.282 |   |   |   |   | 0.299 | 18.5 | 0.0910 | 1.686 | 0.385 | 16.4 | 0.0005 | 0.010 |
| 10 | 0.345 | 17.4 | 0.0613 | 1.066 |   |   |   |   | 0.323 | 17.9 | 0.0755 | 1.354 | 0.385 | 15.4 | 0.0002 | 0.002 |
| 11 | 0.351 | 17.2 | 0.0411 | 0.708 |   |   |   |   | 0.338 | 17.6 | 0.0560 | 0.983 | 0.385 | 15.4 | 0.0001 | 0.001 |
| 12 | 0.358 | 17.1 | 0.0259 | 0.442 |   |   |   |   | 0.347 | 17.3 | 0.0353 | 0.612 | 0.385 | 15.4 | 0.0000 | 0.000 |

Table 1: Error bounds and related statistics for solutions found by exact value iteration and policy iteration in solving the gridworld problem in both its completely and partially observable forms. Only results for the first 12 iterations are shown. In the completely observable case, policy iteration converges after 4 iterations. For each iteration, $\underline{J}$ is the smallest value of any state or belief state, that is, $\underline{J} = \min_{s \in S} J(s)$ in the completely observable case, and $\underline{J} = \inf_b J(b)$ in the partially observable case; $m = (1 - \underline{J})/0.04 + 1$ is an upper bound on the expected number of steps until termination, beginning from *any* state; *resid.* is $||TJ - J||$, the Bellman residual; and *error* $= m \cdot$ *resid.* is the suboptimality bound. Iteration 0 is for the value function of the uniform random policy.

|   | J(0) | N(0) | J(1) | N(1) | J(2) | N(2) | J(3) | N(3) | J(4) | N(4) | J(5) | N(5) | J(6) | N(6) | J(7) | N(7) | J(8) | N(8) | J(9) | N(9) | J(10) | N(10) |
|---|---|---|---|---|---|---|---|---|---|---|---|---|---|---|---|---|---|---|---|---|---|---|
| 0 | -1.28 | 58.0 | -0.88 | 48.0 | -0.32 | 34.0 | 1.00 | 1.0 | -1.52 | 64.1 | -0.92 | 49.0 | -1.00 | 51.0 | -1.60 | 66.1 | -1.52 | 64.1 | -1.28 | 58.1 | -1.22 | 56.5 |
| 1 | 0.81 | 5.7 | 0.87 | 4.3 | 0.92 | 3.1 | 1.00 | 1.0 | 0.76 | 7.0 | 0.66 | 9.5 | -1.00 | 51.0 | 0.68 | 9.1 | 0.39 | 16.3 | 0.44 | 15.0 | -0.88 | 48.1 |
| 2 | 0.81 | 5.7 | 0.87 | 4.3 | 0.92 | 3.1 | 1.00 | 1.0 | 0.76 | 7.0 | 0.66 | 9.5 | -1.00 | 51.0 | 0.71 | 8.4 | 0.66 | 9.6 | 0.59 | 11.2 | 0.37 | 16.8 |
| 3 | 0.81 | 5.7 | 0.87 | 4.3 | 0.92 | 3.1 | 1.00 | 1.0 | 0.76 | 7.0 | 0.66 | 9.5 | -1.00 | 51.0 | 0.71 | 8.4 | 0.66 | 9.6 | 0.61 | 10.7 | 0.39 | 16.3 |
| 4 | 0.81 | 5.7 | 0.87 | 4.3 | 0.92 | 3.1 | 1.00 | 1.0 | 0.76 | 7.0 | 0.66 | 9.5 | -1.00 | 51.0 | 0.71 | 8.4 | 0.66 | 9.6 | 0.61 | 10.7 | 0.39 | 16.3 |

Table 2: State values $J(i)$ and upper bounds $N(i)$ on the expected number of steps until termination for each state $i$ of the completely observable gridworld when solved by policy iteration, beginning from the uniform random policy.

The bounds shown in Table 1 are based on Equation 14, which means that they are independent of the starting state. But Equation 13 (upon which Equation 14 is based) lets us compute a separate suboptimality bound for each state. These bounds are proportional to the expected number of transitions needed to reach the goal from each state. In this respect, they are more realistic than the well-known suboptimality bound for discounted infinite-horizon MDPs, which is $||J^* - J|| \leq ||TJ - J||/(1-\beta)$. The latter assumes that the expected number of transitions until termination is $1/(1-\beta)$, regardless of the starting state and its distance from the goal. A second well-known disadvantage of the suboptimality bound for discounted infinite-horizon MDPs is that it is looser, and converges more slowly, the closer the discount factor is to 1. Interestingly, there is something analogous for stochastic shortest path problems. The suboptimality bound of Equation 14 is looser, and converges more slowly, the greater the expected number of transitions until termination starting from the "farthest" state from the goal, that is, the greater $\max_{i \in S} N(i)$. An advantage of using Equation 13 to compute a separate suboptimality bound for each state is that many or most bounds can be relatively tight and converge quickly, even if there is some state that is very far from the goal state with a suboptimality bound that converges more slowly.

For the gridworld example in its completely observable form, Table 2 shows the value $J(i)$ and upper bound $N(i)$ for each state $i$, for each iteration of policy iteration. The smaller $N(i)$ is, the tighter the suboptimality bound. Because the values of states close to the goal tend to converge faster than the values of states farther from the goal, as illustrated in Table 2, their suboptimality bounds also converge faster. Table 2 does show one anomaly. The upper bound $N(6)$, which is for the $-1$ state, is unrealistically high because it ignores the fact that taking any action in this state causes an immediate transition to the terminal state. Taking into account the transition probabilities for this problem, we could set $N(6)$ equal to 1. Thus we ignore it when computing the bound of Equation 14.

The partially observable version of this problem does not have a simple solution. After 12 iterations, the number of vectors used to represent the value function is more than $10,000$! This underscores that, in the partially observable case, suboptimality bounds are especially useful. The complexity of each iteration of policy or value iteration can grow exponentially in the number of iterations, and it is usually not possible to compute an optimal solution.

# 5 General case

We next consider the general case in which there are no restrictions on transition costs and not all policies are proper. In this case, it is no longer possible to use the minimum cost of a transition to bound the average number of stages until termination of a policy $\mu$ for which $J_\mu \leq J$, since the minimum cost could be zero or negative. The analysis needs to be more complex. Although we are not yet able to describe a good approach to computing suboptimality bounds in the general case, we present some preliminary results in this direction. First we show that the dynamic programming operator for stochastic shortest path problems behaves like an $m$-stage contraction operator, even in the general case. We also show that it is possible to compute suboptimality bounds in the general case, although the bounds we describe are much too loose to be practical. Finally, we discuss some implications of these results.

## 5.1 Contraction property

We begin by showing that the dynamic programming operator behaves like an $m$-stage contraction when applied to a value function $J$ that is uniformly improvable. This result is related to Theorem 2, which shows that a greedy policy with respect to a uniformly improvable value function is a proper policy. For any proper policy, there exists (by Definition 1) a finite positive integer $m$ such that there is a positive probability of reaching the terminal state after at most $m$ stages when following this policy, regardless of the start state. In the special case of positive transition costs, Theorem 5 shows how to compute such an $m$; in that case, $m = \max_{i \in S} N(i)$. We now show how to compute such an $m$ for the general case in which there are no restrictions on transition costs.

Consider a finite-horizon MDP that has the same state set, action set, transition probabilities and transition costs as the original stochastic shortest path problem, but where the objective is to compute the minimum $k$-stage cost, $J_k(i)$, for each starting state $i$, of any policy that does *not* reach the terminal state within $k$ stages. In solving this finite-horizon MDP, we must partition the state set into two subsets that need to be treated separately at each stage $k$. For one set of states, denoted $T_k$, termination within $k$ stages is inevitable under *any* policy. For the remaining states, $S \setminus T_k$, there is some policy under which the probability of termination within $k$ stages is zero. It is only for the second set of states that we solve the finite-horizon MDP for $k$ stages. Note that $T_k$ is a proper subset of $S$ for all $k$, unless all policies are proper.

Figure 2 gives the pseudocode for an algorithm that solves this finite-horizon MDP while distinguishing between these two sets of states. The horizon is determined dynamically.

```
1   k := 0    /* k is stage */
2   ∀i ∈ S : J_k(i) := 0, a(i) := ∅   /* initialization */
3   T_k := {t}   /* termination within k stages is inevitable */
4   while (∃i ∈ S\T_k such that J_k(i) ≤ J̃(i))
5       k := k + 1
6       T_k := T_{k-1}
7       for each ( i ∈ S\T_k )
8           a(i) := {u ∈ U : ∑_{j ∈ T_k} p_{ij}(u) = 0}
9           /* a(i) is set of actions that don't lead to a state in T_k */
10          if (a(i) = ∅) then /* all actions lead to termination */
11              T_k := T_k ∪ {i}
12          else /* compute minimal k-stage cost for this state */
13              J_k(i) := min_{u ∈ a(i)} ∑_j p_{ij}(u)(g(i,u,j) + J_{k-1}(j))
```

Figure 2: Algorithm for solving finite-horizon MDP in order to compute a bound $m$ on the number of stages before termination with some positive probability of any policy $\mu$ for which $J_\mu \leq J$.

It is the smallest stage $k$ for which $J_k(i) + a > J(i), \forall i \in S \setminus T_k$, where $a = \min\{g(i, u, t) : i \in S, u \in U\}$ denotes the minimum cost of any transition to the terminal state and $J$ is a uniformly improvable value function for the original stochastic shortest path problem. We know the horizon must be finite because $a$ and $J(i)$ are finite, and (assuming there exists a policy under which termination can be delayed indefinitely beginning from state $i$), $J_k(i)$ goes to infinity as $k$ goes to infinity, by Assumption 3. Since $J_k(i)$ is the minimum cost of any policy that does not reach the terminal state within $k$ stages beginning from state $i$, $J_k(i) + a$ is a lower bound on the cost of any policy that reaches the terminal state in $k + 1$ stages beginning from state $i$. The key observation is that for any policy $\mu$ that does not reach the terminal state within $k + 1$ stages with probability greater than zero beginning from state $i$, we have $J_\mu(i) \geq J_k(i) + a > J(i)$, which contradicts the assumption that $J_\mu \leq J$. Therefore, we can set $m$ equal to one plus the smallest $k$ for which $J_k(i) + a > J(i), \forall i \in S \setminus T_k$, and we have the following result.

**Theorem 6.** *For any stochastic shortest path problem and for any initial value function $J$ for which $TJ \leq J$, the dynamic programming operator $T$ behaves like an $m$-stage contraction operator.*

We say that the dynamic programming operator *behaves like* an $m$-stage contraction operator, not that it is one, because Bertsekas and Tsitsiklis [3, 4] give an example that shows that the dynamic programming operator for stochastic shortest path problems is not a contraction with respect to any norm, unless all policies are proper. (The example is briefly reviewed in Section 2.2). The distinction between being an $m$-stage contraction and behaving like one is necessary if we adopt the definition that an operator is an $m$-stage contraction if and only if it satisfies the condition expressed by Equation 7 for *all* bounded value functions [2]. But Theorem 6 only claims that the dynamic programming operator satisfies the $m$-stage contraction property expressed by Equation 7 for the closed subspace of uniformly improvable value functions. Thus we could also say that the dynamic programming operator for stochastic shortest path problems is an $m$-stage contraction operator in the subspace of uniformly improvable value functions, but not in the space of all bounded value functions.

The question of whether the dynamic programming operator is, or behaves like, a contraction operator is closely related to the possibility of bounding the suboptimality of solutions found by dynamic programming. We can use Equations 13 and 14 to compute suboptimality bounds only if we can bound the average number of stages until the terminal state is reached, and we can bound the average number of stages until the terminal state is reached only if there is some finite $m$ such that policy execution terminates within $m$ stages with positive probability. Conversely, the existence of some finite $m$ such that policy execution terminates within $m$ stages with positive probability implies that we can bound the average number of stages until termination of any policy. But a bound on the number of stages of policy execution required to reach the terminal state with positive probability is not also a bound on the *average* number of stages it takes to reach the terminal state. The second bound is at least as great as the first, but it is usually greater.

Given that policy execution terminates within $m$ stages with positive probability, where $m$ is computed as above, we can describe a simple approach to bounding the average number of stages it takes to reach the terminal state. Let $p_t = \min\{p_{it}(u) > 0 : i \in S, u \in U\}$ denote the smallest non-zero probability of a transition into the terminal state $t$ and let $p_n = \min\{p_{ij}(u) > 0 : i \in S, j \in S \backslash t, u \in U\}$ denote the smallest non-zero probability of any other transition. It follows that $\rho_m = p_n^{m-1} \cdot p_t > 0$ is a lower bound on the probability of process termination within $m$ stages. Thus an upper bound on the expected number of stages until termination is given by $m \sum_{t=0}^{\infty} (1 - \rho_m)^t = m/\rho_m$. If we set $N(i) = m/\rho_m, \forall i \in S$, we can use Equations 13 and 14 to compute suboptimality bounds. But almost surely, these bounds will be much too loose to be of any practical value. Their derivation does show that it is possible to use the Bellman residual to compute suboptimality bounds, however, and future work may lead to a more sophisticated approach that computes tighter bounds.

## 5.2 Implications

Regardless of whether Theorem 6 supports a practical approach to computing suboptimality bounds in the general case, it has some important theoretical implications. Among them, it points to a stronger convergence proof for policy iteration than the proofs given by Bertsekas and Tsitsiklis [3, 4] in the completely observable case, and by Patek [11] in the partially observable case. Because policy iteration must start with an initial proper policy, the dynamic programming operator used in the policy improvement step behaves like an $m$-stage contraction, and thus standard contraction theory can be invoked to establish uniform convergence. The $m$-stage contraction behavior also establishes that policy iteration converges at a geometric rate. In addition, it establishes that value iteration converges at a geometric rate when given an initial value function that is uniformly improvable. By contrast, the convergence proofs of Bertakas and Tsitsiklis [4] and Patek [11] do not establish that policy iteration and value iteration converge at a geometric rate, unless all policies are proper.

The significance of ensuring a geometric rate of convergence is that for any $\epsilon > 0$, it is possible to bound the number of iterations of value (or policy) iteration needed to find an $\epsilon$-optimal policy. In the partially observable case, this result is especially noteworthy. The problem of finding an $\epsilon$-optimal policy for a discounted infinite-horizon POMDP is well-known to be decidable, by the contraction property of the dynamic programming operator in the case of discounting. But for undiscounted infinite-horizon POMDPs, the problem of finding an $\epsilon$-optimal policy has been shown to be undecidable, in general [7]. However, our results imply that for an important special case of undiscounted infinite-horizon POMDPs, the partially observable stochastic shortest path problem, the problem of finding an $\epsilon$-optimal policy is decidable.

To help make this result seem more plausible, note that the undecidability of $\epsilon$-approximation for undiscounted infinite-horizon POMDPs is proved by reduction from the problem of maximizing the probability of reaching a goal state, where there is a reward of 1 for reaching the goal state, a reward of 0 for not reaching the goal state, and the goal state cannot be reached with probability 1 [7]. Obviously, this problem cannot be reduced to a partially observable stochastic shortest path problem. On the other hand, the optimization problem for discounted infinite-horizon POMDPs, which is also undecidable [7], *can* be reduced to a partially observable stochastic shortest path problem. But it does not imply the undecidability of $\epsilon$-approximation, since $\epsilon$-approximation is decidable for discounted infinite-horizon POMDPs.

Two key assumptions of the stochastic shortest path problem play a role in making $\epsilon$-approximation decidable in the partially observable case; (i) a proper policy exists, and (ii) the expected cost of policy execution beginning from any state from which the terminal state is not reached with probability 1 is infinite. Combined with our observation that $J_\mu \leq J$ for any policy $\mu$ found by dynamic programming when the initial value function $J$ is uniformly improvable, we have been able to establish the $m$-stage contraction behavior of the dynamic programming operator in the space of uniformly improvable value functions. In turn, it allows use of the Bellman residual of the dynamic programming operator to compute suboptimality bounds.

## 6 Conclusion

For stochastic shortest path problems, we have shown that under the condition that the initial value function is uniformly improvable, a greedy policy with respect to any value function found by value iteration is proper. We have also shown how to bound the expected number of stages before the terminal state is reached when following a proper policy found by either value iteration or policy iteration, which in turn lets us use the Bellman residual of the dynamic programming operator to compute suboptimality bounds. The key formula used to compute suboptimality bounds is due to Bertsekas [2]. But it has not been clear how it could be applied to the case where *not* all policies are proper. Our contribution is to show that it can be used to compute suboptimality bounds even when not all policies are proper, as long as the initial value function is uniformly improvable.

In the special case of positive transition costs, especially when the transition costs are uniform or nearly-uniform, we showed that useful suboptimality bounds can be easily computed. In the general case in which transition costs can be zero or negative, we showed that suboptimality bounds are possible, but without describing a practical approach to computing bounds that are tight enough to be useful. We leave this problem for future work.